\documentclass{article}

\usepackage{PRIMEarxiv}

\usepackage[utf8]{inputenc} % allow utf-8 input
\usepackage[T1]{fontenc}    % use 8-bit T1 fonts
\usepackage{hyperref}       % hyperlinks
\usepackage{url}            % simple URL typesetting
\usepackage{booktabs}       % professional-quality tables
\usepackage{amsfonts}       % blackboard math symbols
\usepackage{nicefrac}       % compact symbols for 1/2, etc.
\usepackage{microtype}      % microtypography
\usepackage{lipsum}
\usepackage{fancyhdr}       % header
\usepackage{graphicx}       % graphics
\usepackage{listings}
\usepackage{amsmath}
\usepackage{chngcntr}

\counterwithout{table}{section}   % 표는 섹션번호 떼기
\graphicspath{{/}}     % organize your images and other figures under media/ folder

\title{Scaling Up without Fading Out: Goal-Aware Sparse GNN for RL-based Generalized Planning

}

\author{
  Sangwoo Jeon, Juchul Shin, Gyeong-Tae Kim, YeonJe Cho and Seongwoo Kim \\
  Unmanned Ground Control Technology Lab \\
  LIG Nex1\\
  Daejeon\\
  \texttt{\{sangwoo.jeon\}@lignex1.com} \\
}

\begin{document}
\maketitle

\begin{abstract}
Generalized planning using deep reinforcement learning (RL) combined with graph neural networks (GNNs) has shown promising results in various symbolic planning domains described by PDDL. However, existing approaches typically represent planning states as fully connected graphs, leading to a combinatorial explosion in edge information and substantial sparsity as problem scales grow, especially evident in large grid-based environments. This dense representation results in diluted node-level information, exponentially increases memory requirements, and ultimately makes learning infeasible for larger-scale problems. To address these challenges, we propose a sparse, goal-aware GNN representation that selectively encodes relevant local relationships and explicitly integrates spatial features related to the goal. We validate our approach by designing novel drone mission scenarios based on PDDL within a grid world, effectively simulating realistic mission execution environments. Our experimental results demonstrate that our method scales effectively to larger grid sizes previously infeasible with dense graph representations and substantially improves policy generalization and success rates. Our findings provide a practical foundation for addressing realistic, large-scale generalized planning tasks.
\end{abstract}

% keywords can be removed

\section{Introduction}

Autonomous drones operating in real-world environments must not only navigate efficiently but also respect complex domain constraints (e.g., safety zones, no-fly areas, reconnaissance priorities). Symbolic AI, and in particular PDDL-based planning, offers a principled way to encode such rich domain knowledge and safety requirements directly into the decision-making pipeline\cite{mcdermott1998pddl}\cite{yang2022safe}.  We aim to harness these advantages by integrating PDDL planning into an RL framework for drone mission planning, whereby learned policies inherit the expressiveness and verifiability of symbolic models. For instance, safety zones and no-fly areas can be encoded as hard constraints in PDDL, ensuring certified adherence during both training and deployment.

However, most existing approaches construct fully connected graphs among all objects, which leads to severe scalability bottlenecks in large-scale environments\cite{gupta2024graphscale}. In grid-based domains, such as navigation or mission planning scenarios, this dense connectivity causes exponential growth in edge count, massive sparsity in meaningful features, memory explosion, and ultimately failure to train reinforcement learning agents effectively. These limitations critically hinder the application of generalized planning to realistic, large-area problems.

Moreover, drone missions often involve multi-stage objectives—navigation, target scanning, emergency rerouting—demanding that learned planners handle complex tasks beyond simple waypoint following\cite{hayat2020multi}.  Thus, we also need to demonstrate that our approach not only scales up but also generalizes to these richer, multi-objective missions.

In this paper, we propose a scalable, goal-aware sparse graph representation that selectively encodes only local adjacency relationships while explicitly embedding spatial features relative to the goal. By preserving meaningful structure and injecting goal-directed signals into node features, our method substantially improves both training efficiency and policy generalization capability in large grid-based environments. We validate our approach by designing and solving novel PDDL problems modeled after drone mission planning in expansive grid worlds, demonstrating successful scalability to grid sizes beyond what was previously feasible.

Our results suggest a practical starting point for extending generalized planning techniques to realistic, high-dimensional settings without suffering from representation bottlenecks or information dilution, offering new opportunities for applying learned planners to real-world autonomous systems.

Our contributions are summarized as follows:

\begin{itemize}
\item We propose a sparse, goal-aware graph representation that enables scalable planning.
\item We demonstrate the first successful RL-based generalized planning on large grid-based domains defined by PDDL.
\item We validate our approach on drone-world PDDL problems, showing significant improvements in success rate and generalization.
\end{itemize}

\section{Related Works}

\subsection{Planning Domain Definition Language (PDDL)}
\label{sec:pddl}

The Planning Domain Definition Language (PDDL) is a widely used standard for expressing planning problems in a symbolic and structured manner\cite{mcdermott1998pddl}. It separates domain knowledge—such as action models, object types, and predicates—from specific problem instances that define the initial and goal states. This abstraction enables planners to solve multiple instances with shared dynamics and has become central in generalized planning research.

A representative example is the \textit{Blocksworld} domain, where the environment consists of blocks that can be stacked and unstacked under symbolic constraints. The domain is defined using first-order predicates and typed actions. For instance:

% listings 전역 설정
\lstset{
  basicstyle=\ttfamily\scriptsize, % 1컬럼에서 너무 커보이면 \scriptsize 추천
  columns=flexible,                % fullflexible 대신 flexible로 간격 깨짐 방지
  keepspaces=true,                 % 공백 유지 (정렬 깨짐 방지)
  breaklines=true,                 % 자동 줄바꿈
  breakatwhitespace=false,
  frame=single,
  showstringspaces=false,
  upquote=true,
  tabsize=2,
  xleftmargin=0pt,
  framexleftmargin=0pt
}

% (선택) PDDL/Lisp 키워드 하이라이트를 원하면 간단 언어 정의
\lstdefinelanguage{PDDL}{
  morekeywords={:predicates,:action,:parameters,:precondition,:effect,and,not},
  sensitive=true
}
\begin{lstlisting}[language=PDDL, linewidth=\columnwidth]
(:predicates
  (on ?x - block ?y - block)
  (clear ?x - block)
  (handempty)
  (holding ?x - block))

(:action stack
 :parameters 
   (?x - block ?y - block)
 :precondition 
   (and (clear ?x) 
        (clear ?y) 
        (holding ?x))
 :effect 
   (and (not (clear ?y)) 
        (not (holding ?x))
        (on ?x ?y) 
        (clear ?x) 
        (handempty)))

(:action unstack
  ...
)
\end{lstlisting}

This example shows how symbolic behavior is modeled through predicates like \texttt{(on ?x ?y)} and actions such as \texttt{stack}, with clearly defined preconditions and effects. Such compact symbolic modeling supports planning across variable-sized problem instances.

Once the domain is defined, diverse problems can be generated by varying the initial conditions and goals. This makes PDDL especially suitable for generalized planning and learning transferable policies. An example problem instance derived from the above domain is provided in Appendix~\ref{appendix:blocksworld-problem}.

\vspace{1em}
\noindent \textbf{Key PDDL components}:
\begin{itemize}
\item \textbf{Domain:} The abstract environment containing types, predicates, and actions.
\item \textbf{Problem:} A specific instance specifying the objects, initial state, and goals.
\item \textbf{Predicate:} A symbolic statement (i.e., a logical condition or relation) over objects, such as \texttt{(on ?x ?y)}, that describes the state of the world.
\item \textbf{Action:} A symbolic operator consisting of parameters, preconditions, and effects that define how states can change.
\item \textbf{Precondition:} Conditions that must hold for an action to be executed.
\item \textbf{Effect:} Conditions that change after the action is executed.
\end{itemize}

\subsection{Generalized Planning with Deep Reinforcement Learning}
Traditional approaches to generalized planning relied on symbolic reasoning and domain-specific heuristics. However, recent work  introduced a learning-based framework for generalized planning using deep reinforcement learning (RL)\cite{rivlin2020generalized}. In this paradigm, the agent is trained to generalize across multiple PDDL problem instances by learning value functions or policies that capture shared structure. Their framework demonstrated that it is possible to approximate generalized solutions using neural representations, setting a precedent for combining symbolic structures with RL.

\begin{figure}[t]
\centering
\includegraphics[width=0.8\columnwidth]{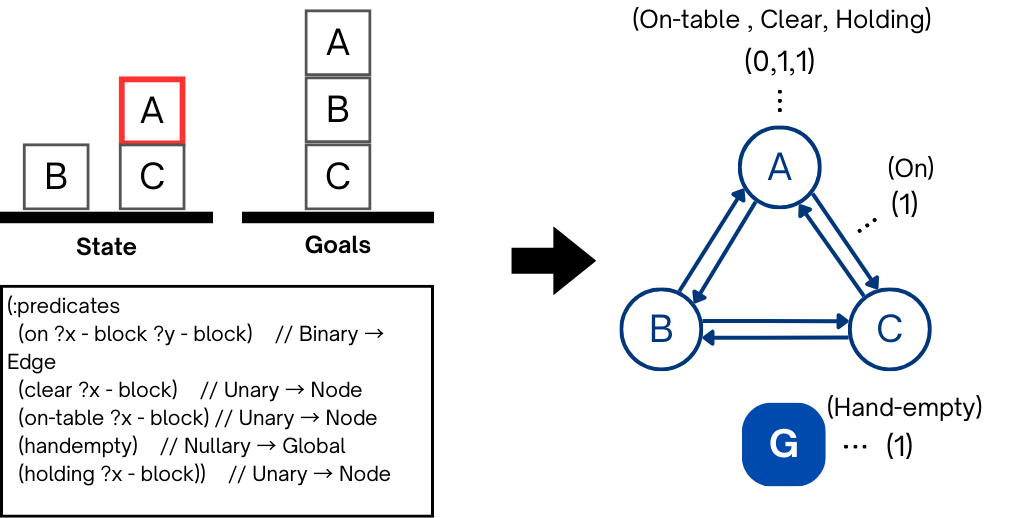} % Reduce the figure size so that it is slightly narrower than the column. Don't use precise values for figure width.This setup will avoid overfull boxes.
\caption{Illustration of the symbolic-to-graph transformation process for the BlocksWorld environment. The environment state (left) shows blocks A, B, and C in a specific configuration (current state), while the goal state (right) defines the target stack arrangement. This state is encoded as a graph (right), where nodes represent the blocks and the global feature “hand-empty” is represented as Global (G). Edges encode binary predicates such as “on” and “clear,” while node features capture unary predicates like “on-table” or “holding.” The resulting graph structure provides a compact and expressive representation of symbolic planning states, serving as input to the GNN-based reinforcement learning pipeline.}
\label{fig1}
\end{figure}

\subsubsection{Reinforcement Learning with Graph Embedding}

Building on this foundation, graph neural networks (GNNs) have emerged as a powerful tool for encoding symbolic planning states in deep reinforcement learning (RL)\cite{janisch2020symbolic}. In this approach, symbolic entities (e.g., blocks, objects) are modeled as graph nodes, and relationships such as binary predicates (e.g., \texttt{(on A B)}) are encoded as directed edges between nodes.

Unary predicates (e.g., \texttt{(clear A)}, \texttt{(on-table B)}) are represented as node features, and nullary predicates (e.g., \texttt{handempty}) are modeled as global features shared across the graph. This graph-based representation, as illustrated in Figure~\ref{fig1}, supports a compact and expressive encoding of planning states that scales to variable numbers of objects.

To model the dynamics of symbolic planning, each environment state is embedded as a graph, and all applicable actions at that state are encoded through their symbolic effects. Each action embedding captures the predicates it adds or deletes, including their scope (global, node, or edge). This information is fused with the corresponding part of the current state graph (e.g., node features for unary effects) to form structured action representations.

The RL agent receives these state and action embeddings and learns a policy $\pi(a|s)$ that maps from symbolic states to the most promising applicable action. During training, the agent is rewarded only upon reaching a goal state (sparse reward setting), while optionally using heuristic-based shaped rewards to stabilize learning in large symbolic spaces.

Through message passing across nodes and edges, GNNs perform relational reasoning over symbolic structure, enabling generalization across varying object configurations. The entire process—symbolic grounding, GNN-based encoding, and RL-based policy learning—forms a unified pipeline for scalable generalized planning in PDDL domains.

\subsubsection{Action Embedding and Selection}

Applicable actions are generated through symbolic grounding using a PDDL parser and engine, extracting all feasible actions based on the current state. Each action is embedded based on its symbolic effects, capturing added and deleted predicates across global, node-level, and edge-level states. These action embeddings are explicitly aligned with corresponding state features, enabling structured representations:

\begin{equation}
\text{score}(a, s) = \text{MLP}([\mathbf{e}_a; \mathbf{x}_s])
\end{equation}

where $\mathbf{e}_a$ denotes the symbolic action embedding, and $\mathbf{x}_s$ refers to relevant state embeddings from affected nodes and edges.

The action selection policy computes a probability distribution over applicable actions through a GNN-based policy network, typically using a softmax function:

\begin{equation}
\pi(a|s) = \frac{\exp(\text{score}(a, s))}{\sum_{a'} \exp(\text{score}(a', s))}
\end{equation}

The highest-scoring action is executed, updating the graph representation accordingly and forming a closed-loop reinforcement learning cycle.

\subsubsection{Scalability Limitations in Grid-Based Domains}
Despite their advantages, existing GNN-based methods often rely on fully connected graphs, connecting all object pairs regardless of their spatial or logical relevance. In grid-based domains, this leads to a combinatorial explosion in the number of edges—resulting in severe memory consumption, increased computational cost, and diluted learning signals due to feature sparsity\cite{gupta2024graphscale}. These limitations render dense representations infeasible for large-scale problems, such as drone navigation over $15\times15$ grids or higher.

\subsection{Techniques to Address Grid-Based Scalability}

\subsubsection{Sparse Graph Connectivity}
Sparse connectivity in graph representations significantly mitigates scalability issues. Methods like Value Iteration Networks (VIN) leverage local interactions, employing localized connections and parameter sharing to control memory and computational complexity\cite{wang2024highway}. Recent advancements, such as Dynamic Transition VIN, effectively scale to large grids (e.g., $100\times100$) through techniques like adaptive skip connections \cite{wang2024scaling}. Attention mechanisms and multi-hop message passing in GNNs further reduce edge explosion by dynamically focusing on relevant graph subsets \cite{janisch2020symbolic}.

\subsubsection{Hierarchical Abstraction}
Hierarchical planning techniques, such as Hierarchical Path-Finding A* (HPA*) \cite{Botea2004HPAstar}, significantly reduce complexity by partitioning large grids into smaller manageable subproblems. Similarly, hierarchical RL methods, including Feudal Networks \cite{Vezhnevets2017Feudal}, abstract tasks into subgoals or macro-actions, dramatically improving scalability and reducing planning horizon.

\subsubsection{Curriculum Learning}
Curriculum learning gradually increases task complexity, improving scalability by first mastering simpler scenarios before scaling to larger grids\cite{bengio2009curriculum}. Lim demonstrated enhanced RL performance on large grids by progressively expanding the environment size, effectively managing exploration difficulty and reward sparsity\cite{lim2020snake}.

\subsubsection{Goal-Aware Reinforcement Learning Approaches}
Goal-aware methods integrate goal-specific information into RL models to address sparse rewards and enhance generalization. Approaches like VIN \cite{tamar2016value} and UVFA \cite{schaul2015universal} utilize goal descriptors within network architectures, while Hindsight Experience Replay (HER) \cite{andrychowicz2017hindsight} improves sample efficiency by reusing experiences. Our proposed method directly incorporates spatial goal information into sparse GNN representations, enhancing generalization and goal-directed behavior.

\subsubsection{Summary}
Addressing scalability in grid-based domains requires integrated approaches that combine sparse connectivity, hierarchical abstraction, curriculum learning, and goal-aware representations. By leveraging these techniques, research has progressively enabled RL methods and GNNs to generalize effectively to large-scale, realistic environments.

\begin{figure*}[t]
\centering
\includegraphics[width=0.95\textwidth]{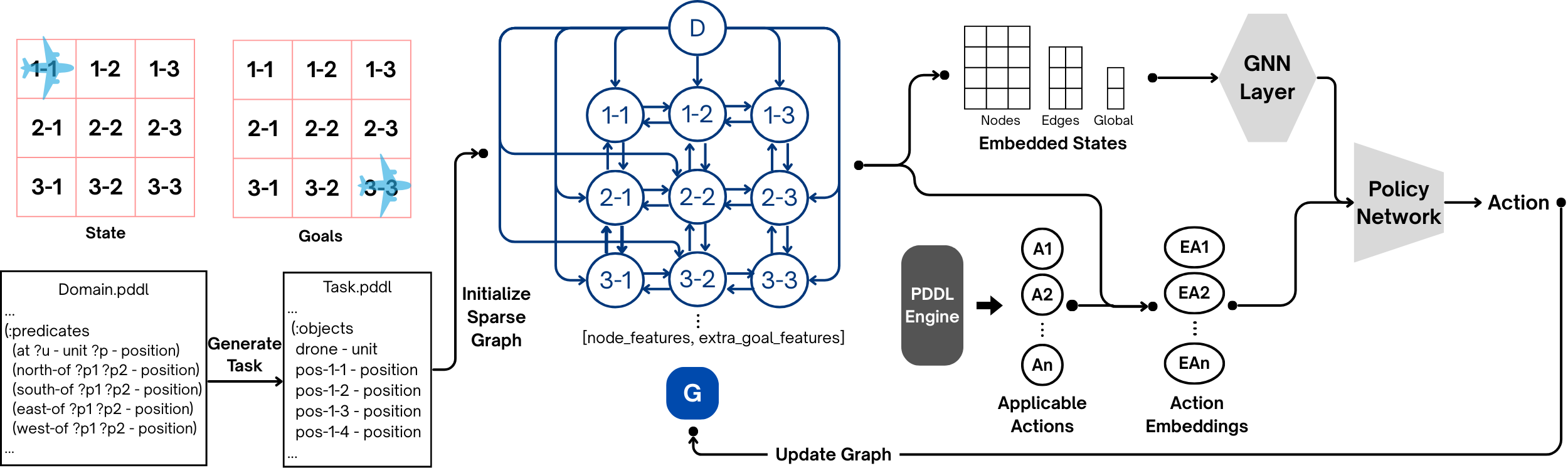} % Reduce the figure size so that it is slightly narrower than the column.
\caption{An overview of the proposed reinforcement learning pipeline for symbolic generalized planning. The environment state and goal are defined in PDDL and converted into a sparse graph representation consisting of nodes (objects), edges (binary predicates), and global features (nullary predicates). Each node embeds both its current state and goal-related features, enabling goal-aware relational reasoning. Applicable actions are extracted by the PDDL engine and encoded through their symbolic effects to form action embeddings. The embedded graph state and action representations are processed by a GNN-based policy network, which outputs the selected action. This action updates the graph, forming a closed reinforcement learning loop.}
\label{fig2}
\end{figure*}

\section{Method}
We now describe our proposed method, a scalable, goal-aware sparse graph representation integrated with reinforcement learning (RL) designed specifically to tackle generalized planning in large-scale symbolic domains such as drone mission planning in grid worlds. As shown in Figure~\ref{fig2}, our proposed method comprises four core components: (1) Sparse Graph Representation, (2) Goal-Aware Node Embedding, (3) Action Embedding and Selection, and (4) Training and Reward Design.

\subsection{Sparse Graph Representation}

We address scalability issues by encoding planning states as sparse graphs rather than fully connected ones. Nodes represent entities such as grid positions, drones, or targets, while edges represent local spatial adjacency or explicitly defined directional relationships (north, east, south, west). Unlike dense graphs, this selective connectivity drastically reduces memory and computational requirements, allowing the method to scale effectively to larger environments.

Edges are created based on spatial adjacency rules explicitly defined through PDDL predicates. For example, grid cells connect only with their four immediate neighbors (up, down, left, right), rather than all other positions in the environment. This leads to a linear number of edges with respect to grid size ( 4N for N positions), compared to quadratic growth in a fully connected graph. Special entities like drones or targets connect selectively to relevant nodes, preserving the meaningful local information essential for efficient graph neural network (GNN) message passing. This sparsification greatly improves scalability in large grid domains.

\subsection{Goal-Aware Node Embedding}

Our node embedding explicitly integrates goal information to enhance the policy's ability to reason about goal-directed actions. Each node’s feature vector includes symbolic state predicates (e.g., occupancy, obstacle presence) and goal-related spatial descriptors, such as relative position, distance, and directional indicators to the goal. Specifically, node embeddings incorporate:

\begin{equation}
\mathbf{x}_i = [\mathbf{s}_i; \mathbf{rel}_i; \mathbf{dist}_i; \mathbf{tgt}_i]
\end{equation}

where $\mathbf{s}_i$ represents symbolic and basic spatial state features, $\mathbf{rel}_i = [\Delta x_i, \Delta y_i]$ encodes the normalized relative offset from node $i$ to the goal, and $\mathbf{dist}_i = [d_i, \theta_i]$ includes the normalized Euclidean distance and the angle (in radians) between node $i$ and the goal direction. For multi-target environments such as droneworld\_scan, $\mathbf{tgt}_i$ consists of target-specific features for each target $t$:

\begin{equation}
\mathbf{tgt}{i,t} = [\text{scanned}t(i), \Delta x{i,t}, \Delta y{i,t}, d_{i,t}, \theta_{i,t}]
\end{equation}

where \texttt{scanned$_t$(i)} is an indicator (1 if scanned at node $i$, 0 otherwise), and the other terms follow the same spatial conventions relative to target $t$. These are concatenated across all targets. This structured embedding significantly boosts generalization capabilities and goal-directed planning effectiveness.

\subsection{Reward Design}

The RL agent undergoes episodic training, executing actions sequentially until achieving the goal or reaching a predefined step limit. We primarily employ a sparse reward structure, granting positive feedback only upon goal completion. 

% --- Distance-Normalized Movement Penalty ---
To discourage inefficient or aimless exploration, we introduce a \emph{distance-normalized movement penalty}. For every drone movement from position \((x_{t-1},y_{t-1})\) to \((x_t,y_t)\) on a grid of size \(W\times W\), we calculate the normalized Euclidean step distance:
\begin{equation}
  d_t = \sqrt{(x_t - x_{t-1})^2 + (y_t - y_{t-1})^2},\quad
  d^{\rm norm}_t = \frac{d_t}{W}.
\end{equation}
A small negative reward is then applied to penalize unnecessary movements:
\begin{equation}
  r^{\rm penalty}_t = -\alpha \cdot d^{\rm norm}_t
\end{equation}
In our experiments, we set \(\alpha=0.1\). This penalty incentivizes more direct, efficient paths toward goal locations.

% --- Sparse Scan Bonus ---
In the multi-stage droneworld\_scan scenario, drones must complete intermediate reconnaissance tasks (scanning targets) before achieving the final goal. To encourage effective task decomposition, the agent receives a sparse bonus whenever a new target is successfully scanned. Formally, if \(S_t\) denotes the set of targets scanned at time step \(t\), the incremental reward is defined as:
\begin{equation}
  r^{\rm scan}_t = 10 \times \bigl|S_t \setminus S_{t-1}\bigr|
\end{equation}
This promotes effective planning toward completing essential subtasks en route to the primary goal.

\subsection{Training and Optimization}

To learn generalized policies over sparse symbolic graphs, we adopt a standard policy gradient framework using Proximal Policy Optimization (PPO) combined with Graph Neural Network (GNN) encoders.

The GNN model is composed of two stacked Graph Network (GN) blocks \cite{battaglia2018relational}, each performing sequential updates on edge, node, and global features. These blocks enable relational reasoning over symbolic structures and goal-aware features, providing a strong inductive bias for generalization.

Policy optimization is carried out via PPO \cite{schulman2017proximal}, an on-policy method known for stable and robust training. It uses clipped surrogate objectives to constrain policy updates within a safe region, improving stability while reusing rollouts for multiple gradient steps.

\subsubsection{Curriculum Learning}

To enhance learning stability and promote better generalization across increasingly complex scenarios, we employ a \emph{curriculum learning} approach. Curriculum learning progressively increases task difficulty during training, enabling agents to first master simpler tasks and subsequently build upon these acquired skills\cite{bengio2009curriculum}.

Specifically, we implement a curriculum based on grid size, gradually scaling up task complexity depending on the agent’s performance. Three primary variables manage this progression:

\begin{itemize}
    \item curriculum\_step: the increment offset added to the base grid size,
    \item success\_in\_row: a counter tracking consecutive episode successes,
    \item curriculum\_threshold: the number of consecutive successes required to increase the grid size.
\end{itemize}

During training, the grid size remains fixed until the agent achieves \(N\) consecutive successful episodes, at which point the curriculum\_step is incremented, success\_in\_row resets to zero, and subsequent episodes are generated with the increased grid size. Conversely, if an episode fails, the success\_in\_row counter resets immediately. This gradual and adaptive process ensures that the agent effectively masters simpler environments before tackling more challenging, larger-scale planning problems, thereby significantly enhancing training efficiency and generalization performance.

\section{Experimental Setup}

\subsection{Environment Descriptions}

To verify the proposed sparse representation and goal-embedding effectiveness in realistic grid environments, we designed comprehensive experiments. Specifically, we aim to confirm: (1) scalability and memory efficiency gains from sparse graph representations, (2) performance enhancements from explicit goal embeddings, and (3) improvements achieved through curriculum learning.

We conducted experiments in two custom-designed drone-world environments defined using Planning Domain Definition Language (PDDL): \textbf{Droneworld\_simple} and \textbf{Droneworld\_scan}. Both simulate realistic drone navigation and reconnaissance tasks within grid-based domains:

\begin{itemize}
\item \textbf{Droneworld\_simple} involves navigation tasks using directional actions (north, east, south, west), constrained explicitly by spatial predicates (e.g., \texttt{north-of}, \texttt{south-of}). 
\item \textbf{Droneworld\_scan} extends Droneworld\_simple by incorporating drone heading states and additional reconnaissance objectives. It introduces scanning actions tracked by predicates such as \texttt{scanned}.
\end{itemize}

\subsection{Training Scenarios}

We generated randomized instances with varying grid sizes (5×5, 10×10, 15×15 and 20×20), obstacle densities, and target placements. Drone initial locations and goals (navigation or reconnaissance) were randomly assigned to assess scalability and generalization. For curriculum learning, tasks progressively increased in complexity from smaller to larger grids (5×5 to 15×15), whereas baseline comparisons utilized randomly sampled grid sizes within this range.

All policies were trained for 100 iterations, each consisting of 100 training episodes and up to 20 gradient update steps. We used a hidden dimension of 512, ReLU activations, a learning rate of 0.0001, a discount factor of 0.99, entropy regularization of 0.01, a clipping ratio of 0.2, and KL-divergence cutoff of 0.01.

\subsection{Performance Metrics}

We evaluated policies using the following metrics:
\begin{itemize}
\item \textbf{Success rate}: Percentage of tasks successfully solved within the step limit.
\item \textbf{Evaluation episode mean reward}: Average undiscounted return during evaluation episodes.
\item \textbf{Plan length}: Average number of steps to achieve the goal, reflecting efficiency.
\item \textbf{Memory usage}: GPU memory used during training (batch size = 200), measured on an NVIDIA RTX 4070 (16 GB).
\item \textbf{Expanded states}: Number of states expanded during inference via GBFS-GNN.
\item \textbf{Runtime}: Total time required to solve evaluation tasks.
\end{itemize}

\subsection{Evaluation Procedure}

We employed two complementary evaluation strategies to assess policy learning progress and generalization capability.

\subsubsection{On-policy Evaluation during Training}
During training, periodic evaluations used greedy rollouts of the current policy. Each evaluation phase consisted of 20 unseen problem instances. The agent executed its policy without exploration, recording average success rate and episode mean reward to monitor learning stability and convergence.

\subsubsection{GBFS-GNN Inference for Generalization}  

After training, we evaluate policies on larger, unseen instances using GBFS-GNN, a greedy best-first search guided by the learned policy and value function\cite{rivlin2020generalized}. The search uses the following heuristic:

\begin{equation}
g(s, a) = \frac{\pi(a|s) \cdot V(s)}{1 + H(\pi(\cdot|s))}
\end{equation}

where \(\pi(a|s)\) is the policy probability, \(V(s)\) is the estimated state value, and \(H(\pi(\cdot|s))\) is the policy entropy.

Intuitively, GBFS-GNN prioritizes actions by combining the learned policy's confidence (\(\pi(a|s)\)) and the estimated state value (\(V(s)\)). This encourages the search to explore actions that the policy considers both promising and valuable. Additionally, the entropy term ((\(H(\pi(\cdot|s))\)) lowers the priority of uncertain actions, thus promoting efficient exploration of the most confident and beneficial paths.

We report evaluation success rate, number of expanded states, and runtime on large instances such as 25×25 grids. These results appear in Figures~\ref{fig7} and ~\ref{fig8}, with Fast Downward (LAMA-first configuration) used as a classical planning baseline\cite{helmert2006fast}.

\section{Results}

\subsection{Effect of Sparse Graph Representation}

\begin{figure}[t]
\centering
\includegraphics[width=0.8\columnwidth]{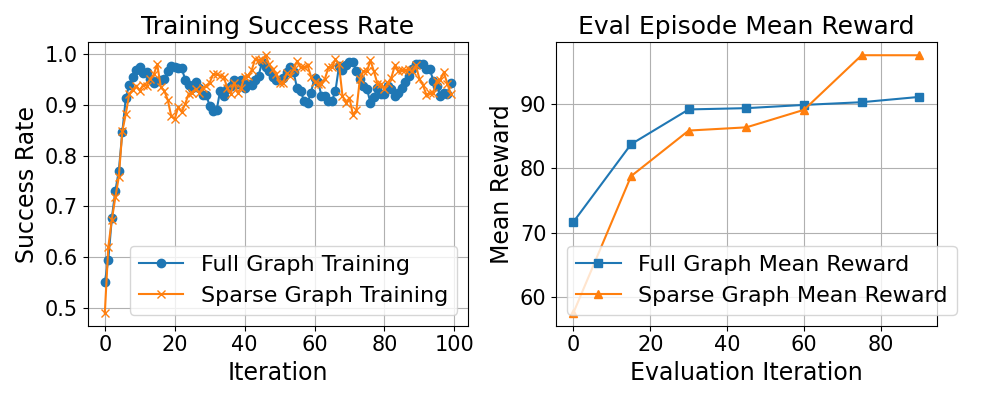} % Reduce the figure size so that it is slightly narrower than the column. Don't use precise values for figure width.This setup will avoid overfull boxes.
\caption{Comparison of Full and Sparse Graph configurations in terms of training success rate (left) and evaluation episode mean reward (right) in the 5×5 grid setting.}
\label{fig3}
\end{figure}

\begin{table}[!t]
\centering
\setlength{\tabcolsep}{4pt}
\caption{Trainability and memory usage comparison between Full Graph and Sparse Graph across various grid sizes.}
\label{table1}
{\footnotesize
\begin{tabular}{lcc}
\hline
\textbf{Grid} & \textbf{Trainable} & \textbf{Memory Usage (GB)} \\ \hline
5$\times$5 Sparse Graph  & Y & 1.3  \\
10$\times$10 Sparse Graph & Y & 3.7  \\
15$\times$15 Sparse Graph & Y & 10.0 \\
20$\times$20 Sparse Graph & Y & 20.0 \\
5$\times$5 Full Graph    & Y & 3.7  \\
10$\times$10 Full Graph  & N & 24   \\
15$\times$15 Full Graph  & N & N/A  \\
20$\times$20 Full Graph  & N & N/A  \\ \hline
\end{tabular}
}
\end{table}

We evaluated the effectiveness of the sparse graph representation in preserving planning performance while reducing computational overhead. As shown in Figure~\ref{fig3}, both sparse and full graph configurations achieved high training success rates in the 5×5 grid environment.

For evaluation performance, the sparse graph configuration not only maintained comparable stability but also achieved a higher episode mean reward in later training iterations (beyond iteration 60). This suggests that sparsity does not hinder learning effectiveness and may even facilitate better policy convergence in practice.
To assess scalability, we measured GPU memory usage and trainability across increasing grid sizes. As summarized in Table~\ref{table1}, the sparse graph representation enabled successful training up to 20×20 grids with significantly lower memory usage.  In contrast, full graph configurations failed to train beyond 10×10 due to excessive memory requirements. All memory measurements were conducted under a fixed batch size of 200. These results confirm that sparse graph representations provide a scalable and efficient alternative without compromising learning performance.

\subsection{Impact of Goal Embedding}

\begin{figure}[t]
\centering
\includegraphics[width=0.7\columnwidth]{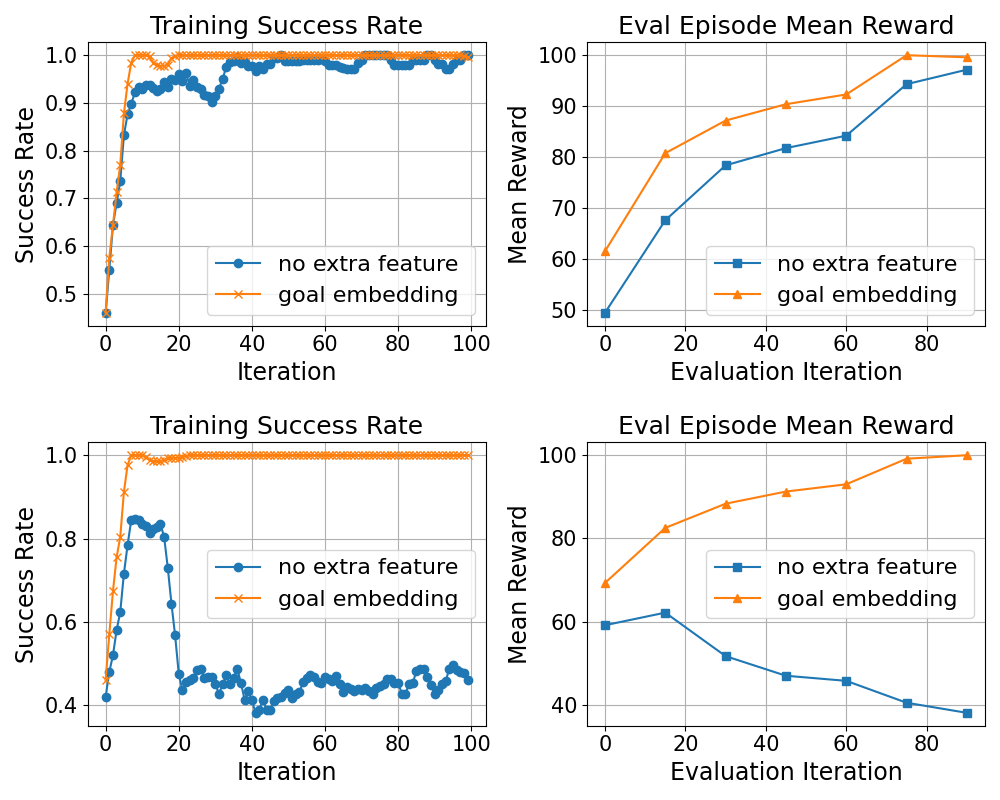}
\caption{Training success rate (left) and evaluation episode mean reward (right) for 5×5 (top row) and 10×10 (bottom row) grids in Droneworld\_simple.}
\label{fig4}
\end{figure}

To evaluate the impact of explicitly embedding goal information into the graph structure, we compared configurations with and without goal embeddings in the \textbf{}{Droneworld\_simple} environment. All experiments were conducted using sparse graph representations for computational efficiency.

As shown in Figure~\ref{fig4}, goal embedding consistently improved learning performance. In the 5×5 grid scenario, the goal-embedded model converged faster and achieved higher evaluation rewards than the baseline. This effect was even more pronounced in the 10×10 setting: the model without goal embedding struggled to learn meaningful policies, exhibiting unstable training success rates and declining evaluation rewards. In contrast, the goal-embedded configuration maintained near-perfect success and continued to improve throughout training.

These results demonstrate that incorporating goal information not only accelerates policy learning but also becomes essential as environment complexity increases, enabling stable and scalable training in larger domains.

\subsection{Effectiveness of Curriculum Learning}

\begin{figure}[t]
\centering
\includegraphics[width=0.8\columnwidth]{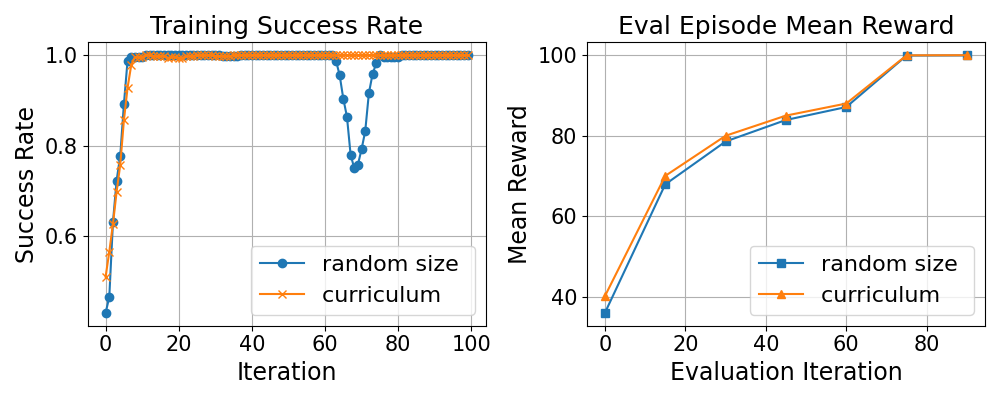}
\caption{Training success rate (left) and evaluation episode mean reward (right) for curriculum learning vs. random-sized training.}
\label{fig5}
\end{figure}

We evaluated the impact of curriculum learning by training agents on progressively larger tasks (from 5×5 up to 15×15) in the \textbf{Droneworld\_simple} environment, and subsequently evaluating them on unseen 20×20 tasks. As shown in Figure~\ref{fig5}, the curriculum-trained model exhibited consistently high success rates and smooth learning dynamics. In contrast, the model trained with randomly sampled grid sizes experienced a noticeable drop in success rate around the middle of training, indicating instability during optimization.

Despite this, both models achieved high evaluation rewards and demonstrated the ability to reach goals efficiently. The average plan length was 21.4 for the randomly trained model and 19.8 for the curriculum-trained model, suggesting both learned near-optimal paths, with curriculum learning yielding slightly more efficient plans.

\begin{figure}[t]
\centering
\includegraphics[width=0.8\columnwidth]{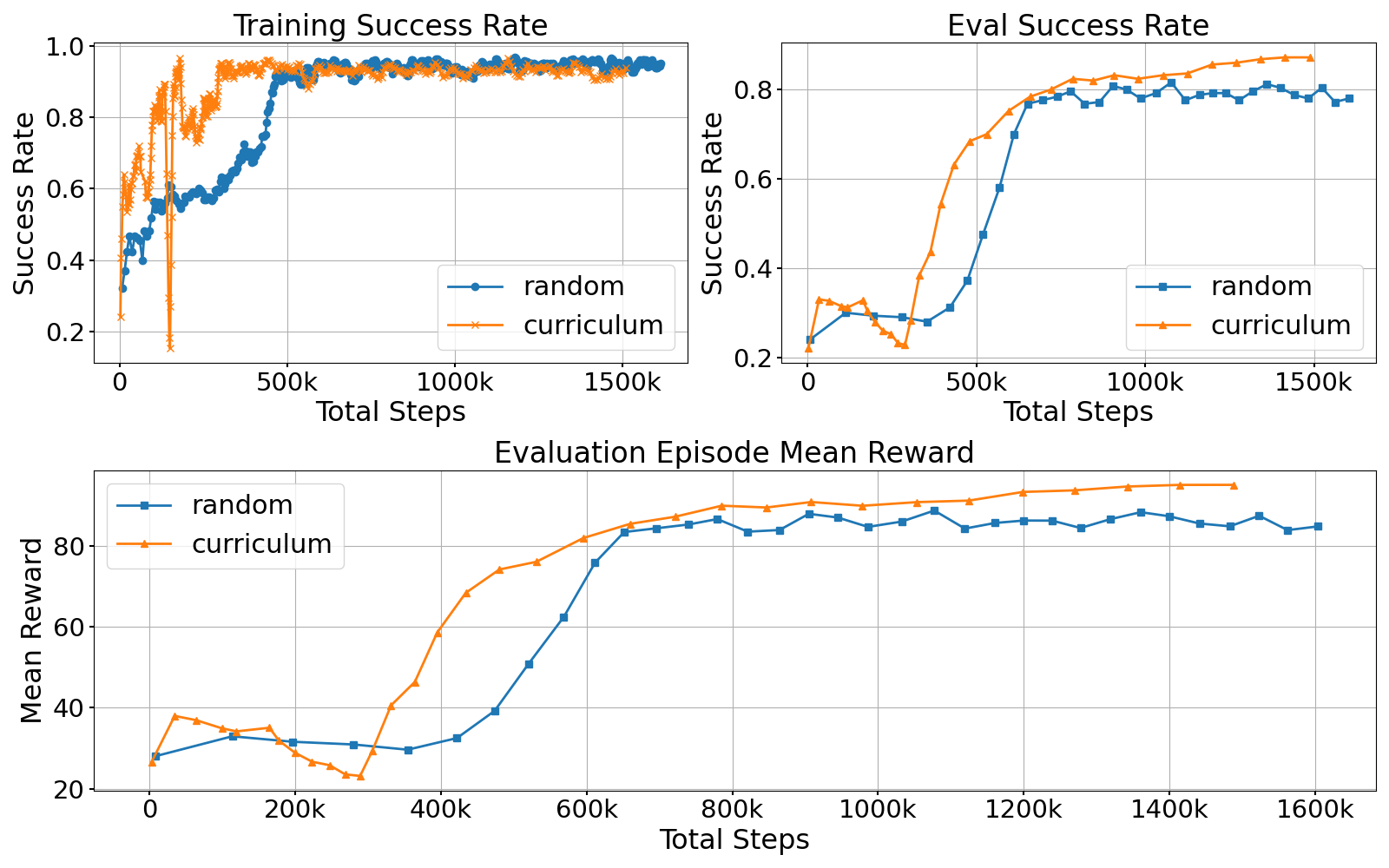}
\caption{Training success rate, evaluation success rate, and evaluation reward at identical step intervals for curriculum versus random-sized training, trained on 5×5 to 12×12 grids and evaluated on unseen 20×20 grids.}
\label{fig6}
\end{figure}

For a more precise and comprehensive comparison, we further conducted experiments using total training steps as a common metric, analyzing training success rate, evaluation success rate, and evaluation rewards, as shown in Figure~\ref{fig6}. Agents were trained on grid sizes ranging from 5×5 to 12×12 and evaluated on unseen 20×20 tasks. The results clearly demonstrate that the curriculum-trained model converged more rapidly during training and consistently achieved higher success rates and rewards during evaluation compared to the randomly trained model.

\begin{figure}[t]
\centering
\includegraphics[width=0.5\columnwidth]{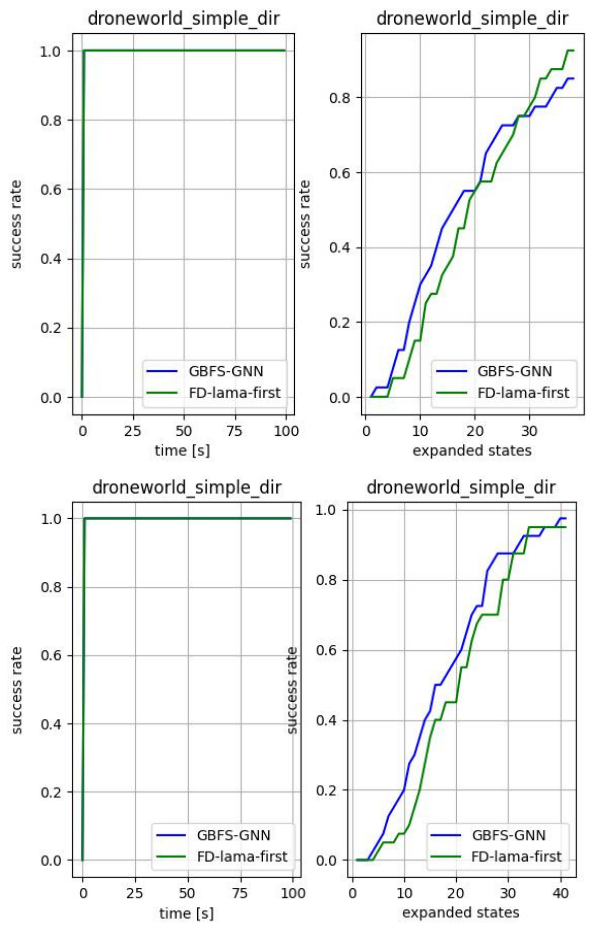}
\caption{Evaluation performance on 25×25 grid: success rate vs. time and expanded states for GBFS-GNN using policies trained with random-sized (top) and curriculum learning (bottom). Fast Downward (FD-lama-first) results are included as a baseline.}
\label{fig7}
\end{figure}

To further assess generalization, we evaluated the learned policies using GBFS-GNN on a larger 25×25 grid. As illustrated in Figure~\ref{fig7}, the curriculum-trained model achieved higher success rates while expanding fewer states compared to the randomly trained counterpart. Although Fast Downward (FD-lama-first) also showed competitive performance, the curriculum-based policy led to more sample-efficient search behavior under the GNN-based planner.

These results demonstrate that curriculum learning enhances training stability and produces policies that generalize better to large, complex environments while maintaining efficient planning performance.

\subsection{Complex Task Generalization}

\begin{figure}[t]
\centering
\includegraphics[width=0.5\columnwidth]{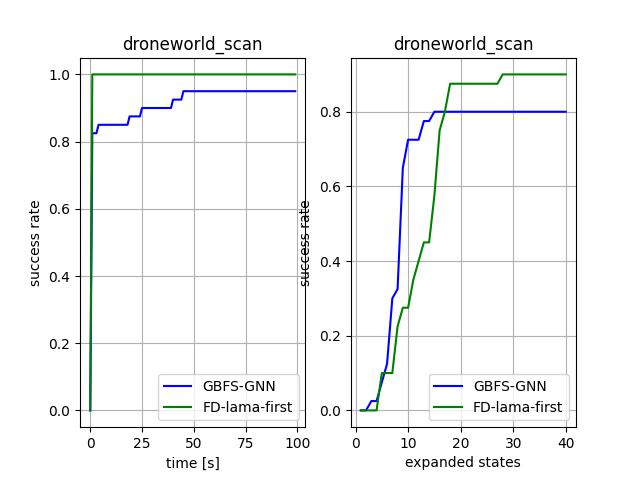}
\caption{Evaluation results in \texttt{Droneworld\_scan} domain at 25×25 grid size using GBFS-GNN. Left: Success rate vs. time. Right: Success rate vs. number of expanded states.}
\label{fig8}
\end{figure}

We further tested the generalization capability of our approach in a more complex scenario using the \textbf{Droneworld\_scan} environment, which requires both navigation and reconnaissance actions. Training was conducted on grids up to size 15×15 using a sparse, goal-aware representation and curriculum learning. The policy achieved perfect training performance, with a success rate reaching 1.0 on 15×15 instances.

Importantly, our experiments demonstrated that curriculum learning led to much more stable training dynamics compared to training with randomly sampled grid sizes. While random grid size training showed noticeable instability and fluctuating performance during learning, the curriculum-based approach provided smooth and consistent improvements across iterations.

To evaluate generalization to larger problems, we deployed the learned policy within the GBFS-GNN search algorithm and tested it on 25×25 instances. As shown in Figure~\ref{fig8}, our method demonstrated high success rates of over 90\%, but slightly lagged behind the classical Fast Downward planner (FD-lama-first) in both runtime and number of expanded states. This indicates that while our learned policy scales well to complex multi-objective domains and benefits significantly from curriculum learning, there remains a performance gap in larger-scale generalization compared to domain-optimized classical planners.

\section{Conclusion and Future Work}

In this paper, we introduced a sparse, goal-aware graph representation combined with curriculum learning to address scalability and generalization challenges in generalized planning tasks using deep reinforcement learning. Our approach demonstrated significant improvements in scalability, substantially reduced GPU memory requirements, and enhanced planning effectiveness compared to traditional dense graph representations. Importantly, our method succeeded in scaling to previously infeasible grid sizes, overcoming fundamental limitations of existing fully-connected graph approaches. Furthermore, goal-aware embeddings and curriculum learning enabled stable training dynamics, significantly improving the robustness and convergence of policy learning.

Through extensive evaluations using two drone mission scenarios, \textbf{Droneworld\_simple} and \textbf{Droneworld\_scan}, we confirmed that progressively complex training via curriculum learning substantially enhances policy generalization and stability. Our results illustrate the advantages of this combined methodology, particularly in larger, realistic mission environments.

For future work, we aim to extend our approach toward practical, real-world drone deployments. Specifically, we plan to integrate the developed generalized planning method into our existing Belief-Desire-Intention (BDI)-based drone framework, enhancing autonomous mission execution with adaptive planning capabilities\cite{Son2024model}. Such integration will enable drones to dynamically adjust their plans in complex, partially observable, and real-time scenarios, combining the symbolic reasoning power of the BDI architecture with the flexibility and scalability provided by our sparse GNN model.

\section*{Acknowledgments}
This research was supported by the Challengeable Future Defense Technology Research and Development Program through the Agency For Defense Development(ADD) funded by the Defense Acquisition Program Administration(DAPA) in 2024(No.912904601)

%Bibliography
\bibliographystyle{unsrt}  
\bibliography{references}  
\relax

\newpage
\section*{Appendix}
\subsection*{Example Blocksworld Problem}
\label{appendix:blocksworld-problem}

Below is an example of a PDDL problem instance corresponding to the Blocksworld domain introduced in Section~\ref{sec:pddl}. It specifies the initial block configuration and the desired goal state. \\

\begin{lstlisting}[language=lisp]
(define (problem bw-prob-1)
  (:domain blocksworld)
  (:objects A B C - block)
  (:init
    (on-table B)
    (on A C)
    (on-table C)
    (clear A)
    (clear B)
  (handempty))
  (:goal
    (and
      (on A B)
      (on B C)))

)
\end{lstlisting}

\subsection*{Droneworld\_simple Domain}
\label{appendix:blocksworld_simple_dir}
\begin{lstlisting}[language=lisp]
(define (domain droneworld_simple_dir)
  (:requirements :typing)
  (:types
    unit
    position
  )
  (:predicates
    (at ?u - unit ?p - position)
    (north-of ?p1 ?p2 - position)  ; p1 is north of p2
    (south-of ?p1 ?p2 - position)
    (east-of ?p1 ?p2 - position)
    (west-of ?p1 ?p2 - position)
  )

  (:action move-north
    :parameters (?u - unit ?from ?to - position)
    :precondition (and (at ?u ?from) (north-of ?to ?from))
    :effect (and (not (at ?u ?from)) (at ?u ?to))
  )

  (:action move-south
    :parameters (?u - unit ?from ?to - position)
    :precondition (and (at ?u ?from) (south-of ?to ?from))
    :effect (and (not (at ?u ?from)) (at ?u ?to))
  )

  (:action move-east
    :parameters (?u - unit ?from ?to - position)
    :precondition (and (at ?u ?from) (east-of ?to ?from))
    :effect (and (not (at ?u ?from)) (at ?u ?to))
  )

  (:action move-west
    :parameters (?u - unit ?from ?to - position)
    :precondition (and (at ?u ?from) (west-of ?to ?from))
    :effect (and (not (at ?u ?from)) (at ?u ?to))
  )
)

\end{lstlisting}

\subsection*{Droneworld\_scan Domain}
\label{appendix:blocksworld_simple_scan}
\begin{lstlisting}[language=lisp]
(define (domain droneworld_scan)
  (:requirements :typing)
(:types
    unit
    position
    direction
)
(:predicates
    (at ?u - unit ?p - position)
    (drone-to ?dir - direction)
    (safe-at ?p - position)
    (adjacent-north ?from ?to - position)
    (adjacent-south ?from ?to - position)
    (adjacent-east ?from ?to - position)
    (adjacent-west ?from ?to - position)
    (scanned ?u - unit)
)

(:action heading-north-forward
    :parameters (?from ?to - position)
    :precondition (and 
                    (at drone ?from)
                    (safe-at ?to)
                    (drone-to north)
                    (adjacent-north ?from ?to)
                  )
    :effect (and 
             (not (at drone ?from))
             (at drone ?to)
            )
)

(:action heading-south-forward
    :parameters (?from ?to - position)
    :precondition (and 
                   (drone-to south)
                   (at drone ?from)
                   (safe-at ?to)
                    (adjacent-south ?from ?to)
                  )
    :effect (and 
             (not (at drone ?from))
             (at drone ?to)
            )
)

...

)(:action scan-west-right
    :parameters (?target - unit ?from ?to - position)
    :precondition (and 
                    (at drone ?from)
                    (drone-to west)
                    (adjacent-north ?from ?to)
                    (at ?target ?to)
    )
    :effect (and 
                (scanned ?target)
    )
)

)

\end{lstlisting}
\end{document}